\documentclass[letterpaper, 10 pt, conference]{ieeeconf} 

\usepackage{amssymb}
\usepackage{cuted}
\usepackage{amsmath}
\usepackage{graphicx}
\usepackage{array}
\usepackage{cite}
\usepackage{multirow}
\usepackage{booktabs}
\usepackage{subfig}
\usepackage{caption}

\IEEEoverridecommandlockouts 

\title{\LARGE \bf
Dual-BEV Nav: Dual-layer BEV-based Heuristic Path Planning \\ for Robotic Navigation in  Unstructured Outdoor Environments
}
\author{
Jianfeng Zhang$^{1*}$, Hanlin Dong$^{1*}$, Jian Yang$^{2*}$, Jiahui Liu$^{3*}$,  Shibo Huang$^{1}$, \\  
  Ke Li$^{2}$,  Xuan Tang$^{1}$, Xian Wei$^{1}$, Xiong You$^{2}$ 
\thanks{$^{1}$Software Engineering Institute, East China Normal University, $^{2}$School of Geospatial Information, Information Engineering University, $^{3}$College of Computer and Cyber Security, Fujian Normal University.
}
}

\begin{document}

\maketitle
\thispagestyle{empty}
\pagestyle{empty}

\begin{abstract}
Path planning with strong environmental adaptability plays a crucial role in robotic navigation in unstructured outdoor environments,
especially in the case of low-quality location and map information.
%
The path planning ability of a robot depends on
the identification of the traversability of global and local ground areas.
In real-world scenarios, the complexity of outdoor open environments makes it difficult for robots to identify the traversability of ground areas that lack a clearly defined structure.
Moreover, most existing methods have rarely analyzed
the integration of local and global traversability identifications in unstructured outdoor scenarios. 
To address this problem,
we propose a novel method, Dual-BEV Nav, first introducing Bird's Eye View (BEV) representations into local planning to generate high-quality traversable paths. Then,
these paths are projected onto the global traversability map generated by the global BEV planning model to obtain the optimal waypoints. 
By integrating the traversability from both local and global BEV, we establish a dual-layer BEV heuristic planning paradigm, enabling long-distance navigation in unstructured outdoor environments. 
We test our approach through both public dataset evaluations and real-world robot deployments, yielding promising results. 
Compared to baselines, the Dual-BEV Nav improved temporal distance prediction accuracy by up to $18.7\%$. 
In the real-world deployment, under conditions significantly different from the training set and with notable occlusions in the global BEV, the Dual-BEV Nav successfully achieved a 65-meter-long outdoor navigation. 
Further analysis demonstrates that the local BEV representation significantly enhances the rationality of the planning, while the global BEV probability map ensures the robustness of the overall planning.

\end{abstract}

\section{INTRODUCTION}

With the rapid advancements in large foundation models~\cite{kim2024survey,rajvanshi2024saynav,wang2024large}, cognitive science~\cite{peller2023memory,valluri2024exploring,newcombe2023building}, and embodied AI~\cite{duan2022survey,song2023llm,sridhar2024nomad}, the path planning of a robot in structured environment has significantly improved. 
Traditional approaches that combine local and global path planning have been reinvigorated by emerging technologies, leading to new possibilities and innovations. 
However, in complex environments characterized by unstructured, open, dynamically changing, path planning for robots still faces serious challenges.

Firstly, in terms of local path planning, the complexity and unstructured nature of outdoor environments make it difficult to identify the traversability of ground areas.
%
To address this challenge, current local path planning primarily adopts the following three kinds of approaches: (i) The manual annotation of ground areas~\cite{thakker2021autonomous}; (ii) Multiple sensor fusion~\cite{weerakoon2023graspe};
(iii) End-to-end navigation via an integrated framework that combines perception, decision-making, and action generation~\cite{shah2021ving,shah2021rapid,shah2023gnm}.
%
However, precise areas' annotation requires a lot of human labor and is difficult to adapt to new environments. Multimodal fusion methods involve substantial computational complexity, making them difficult to deploy on mobile robots. End-to-end navigation approaches impose higher demands on environmental feature extraction.

Secondly, compared to local path planning, global path planning benefits from richer environmental information, offering a foundation for local path decision-making. 
Global path planning relies on prior maps, primarily sourced from (i) Simultaneous Localization and Mapping (SLAM)~\cite{shan2021lvi,shan2020lio} and (ii) Overhead map~\cite{shah2022viking,sanchez2023waypoint}. However, SLAM-based global path planning requires the prior construction of scene maps, but accurate reconstruction in complex unstructured environments is extremely difficult. 
In contrast, prior environmental information (e.g. overhead maps like satellite maps) based on top-down view methods is relatively easy to obtain. However, it suffers from low data resolution and limited information, making it difficult to provide reliable heuristic guidance for local path planning.

To address these issues, we propose a novel method, a Dual-layer BEV-based heuristic path planning framework for robotic Navigation in unstructured outdoor environments (\textit{Dual-BEV Nav}), which consists of two components: (i) Local BEV Planning Model (\textit{LBPM}) and (ii) heuristic Global BEV Planning Model (\textit{GBPM}). 
In \textit{LBPM}, we developed an end-to-end navigation
model based on BEV, extracting traversability features and latent goal features from the BEV representations, which show the advantage of structural integrity and scale invariance.
%
This method allows path planning to have high-accuracy distance calculation, without the explicit drivable area segmentation.
%
%
In \textit{GBPM}, we propose a global BEV model with the capability of identifying traversability hints from the overhead map.
%
Our approach encodes the overhead map into a traversability map by learning from the robot's trajectory. The traversability map is a probability map of hints, where each pixel represents the hint value for the robot to navigate to that location. 
This probabilistic representation significantly reduces the robot's reliance on knowledge of a priori map.
The proposed \textit{LBPM} acts as the controller for the robot’s movements, responsible for generating reliable waypoints to achieve the navigation task. 
The \textit{GBPM} evaluates and selects these waypoints from a global view to identify feasible paths.

The main contributions of this work are as follows:
\begin{itemize}
    \item We propose Dual-BEV Nav, a dual-layer BEV heuristic path planning paradigm for robotic navigation in unstructured outdoor environments. This novel method integrates the traversability from both local and global BEV, completing a 65-meter-long outdoor navigation in an unstructured and challenging environment.
    
    \item We propose a local BEV planning model that can significantly enhance the robot's capability of traversability identification in complex environments, without the need for drivable area segmentation. This makes the robot more adaptable to changes in unknown environments.
    
    \item We propose a global BEV planning model that provides global hints through a probability map, effectively reducing the robot's dependence on precise prior maps for path planning.
\end{itemize}

\begin{figure*}[thpb]
  \centering
  \includegraphics[scale=1.2]{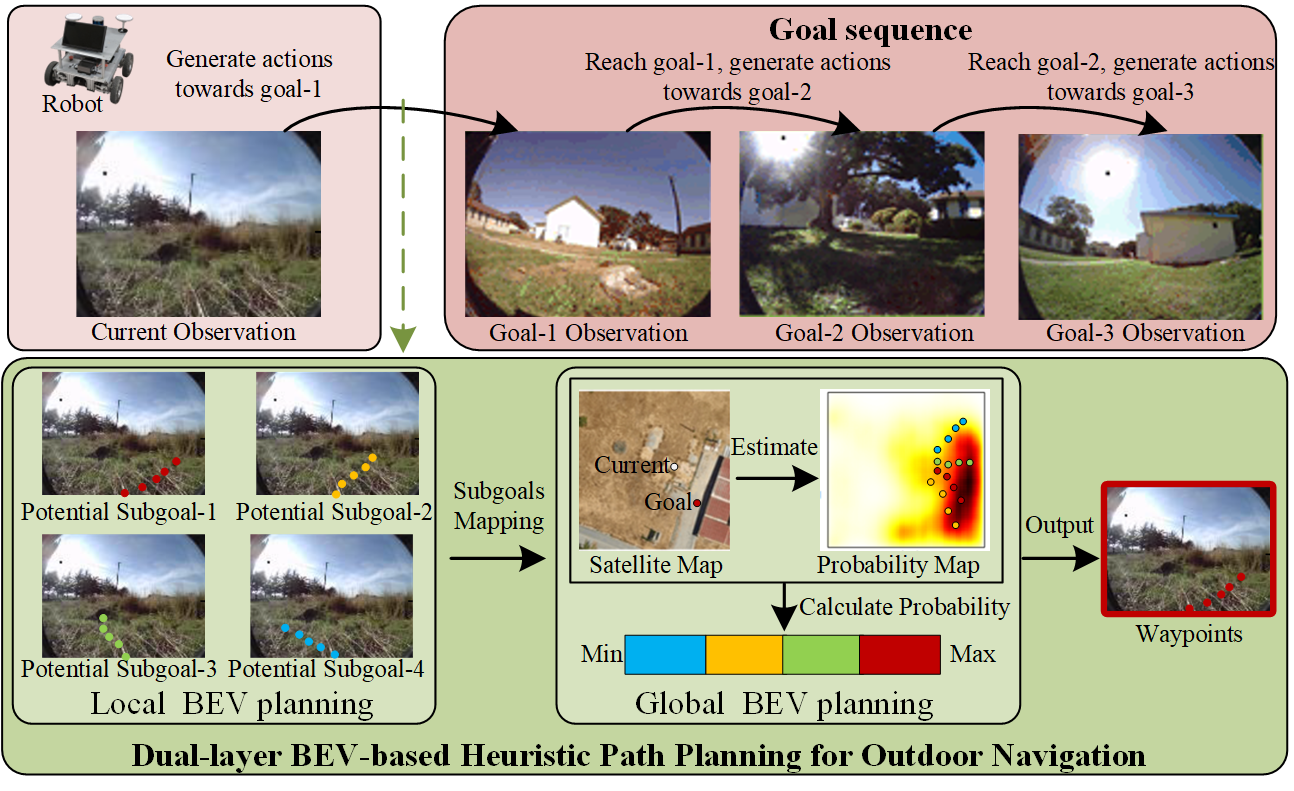}
  \caption{Dual-layer BEV-based heuristic path planning for robotic navigation in unstructured outdoor environments: 
  Dual-BEV Nav integrates traversability information from both local and global BEVs, providing the capability for target exploration and sequential target navigation. The Local BEV planning model identifies the traversability of the local scene. Based on overhead maps, the Global BEV planning model models a probability map of hints. The optimal path is obtained by mapping the local traversable waypoints onto the probability map.}
  \label{Overall architecture}
\end{figure*}

\section{Related Work}
\subsection{The Prior Knowledge of Maps}
The prior knowledge of maps is one of the indispensable pieces of information for robots in global path planning~\cite{delmerico2017active,lluvia2021active,mccormac2017semanticfusion,murthy2019gradslam}, and the acquisition and utilization of this knowledge are key to enabling long-distance path planning. 
Currently, methods that use  knowledge of prior maps for planning can be divided into two categories,
based on the characteristics of the application environment:
(i) Methods for structured environment  ~\cite{yang2019mobile,wang2023dueqnet,brenneke2003using} are widely applied in structured scenarios such as streets, highways, campuses, industrial parks, and ports. These methods leverage distinct structured features to construct precise prior maps. The precise prior map provides a solid foundation for optimizing global paths. However, the heavy reliance on prior maps makes it difficult to reconcile discrepancies between real-time sensor data and prior maps during actual deployment. 
%
(ii) In an unstructured environment, it is difficult to construct precise prior maps.
Consequently, path planning in such environments relies heavily on real-time perception ~\cite{niijima2020city,putz2021continuous,nilwong2019deep}, and the prior knowledge of maps serves as heuristic guidance. These approaches offer greater flexibility and adaptability. However, the limitations of prior maps make it challenging to achieve efficient global planning.

\subsection{The Traversability of Path Planning}
The robot computes path planning through the identification of traversability. 
In the robot's local/global path-planning framework, 
(i) Local traversability identification is based on the robot's real-time perception data. 
However, traditional front-view-based methods inevitably suffer from challenges associated with scale distortions.
In contrast, BEV provides distance information, ensuring scale invariance of the perception. Consequently, BEV-based environmental perception has gained prominence in autonomous driving and outdoor robotics, particularly for tasks such as 3D object detection and semantic segmentation. This technological advancement has also been explored more broadly in the context of path planning ~\cite{hood2017bird,liu2023bird,shaban2022semantic}.
(ii) Global traversability identification relies on the prior knowledge of maps, such as 2D grid maps for A*~\cite{wilt2012does} and Dijkstra algorithms~\cite{noto2000method}, 3D point cloud maps~\cite{wang2021navigation}, and topological maps~\cite{huang2020autonomous}. 
Recently, researchers have focused on identifying traversability from overhead maps, employing techniques such as manual annotation \cite{truong2023indoorsim} and model-based segmentation methods ~\cite{sanchez2023waypoint}. Nonetheless, segmentation approaches relying on color and shape are prone to failure due to substantial geographical variability among regions. Additionally, such approaches often result in significant labor and time costs.

\section{Dual-layer BEV-based Heuristic Path Planning}
As illustrated in Figure \ref{Overall architecture}, the proposed Dual-BEV Nav comprises two components: (i) The \textit{LBPM}, which identifies traversability in the local BEV and generates several traversable waypoints; (ii) The \textit{GBPM}, which decodes the probability map of hints in the global BEV to offer global traversability. The efficient integration of these two models facilitates robotic navigation in unstructured outdoor environments.

\subsection{Local BEV Planning Model}
%

In this work, we propose a novel \textit{LBPM} with excellent capability of identifying traversability, which can adapt to the path planning for robot navigation in unstructured outdoor environments. The model consists of two key components: local BEV perception encoder and task-driven goal decoder. 

\textbf{Local BEV Perception Encoder}: The input to this module consists of two parts: (i) the context observation $o_{t-P:t-1}$ and (ii) the current observation $o_{t}$. Compared to traditional planning methods that rely solely on current observation, this context enables the model to capture environmental changes and learn the robot's motion parameters. 

We introduce the BEV view transformation method to enhance the robot's capability of identifying traversability and adapting to unstructured outdoor environments. 
The BEV representation overcomes common challenges associated with scale distortions that are prevalent in the traditional front-view representation.
Furthermore, it preserves the structural integrity of the 3D scene and provides accurate distance information, significantly improving the accuracy of robot path planning.
Our BEV feature extraction is implemented based on the LSS method~\cite{philion2020lift}. The module utilizes the EfficientNet ~\cite{tan2019efficientnet} to extract image features from the front-view camera.
We follow the methods proposed by LSS~\cite{philion2020lift} and BEVDet~\cite{huang2021bevdet,huang2022bevdet4d} to predict the discrete depth distribution for each pixel. Utilizing the depth estimation $D_i^{estimate}$, we lift the 2D pixel feature $Fea_i^{2D}$ to 3D features $Fea_i^{3D}$ through the following process:
\begin{equation}
    Fea_i^{3D} = D_i^{estimate}Fea_i^{2D}
\end{equation}

We project a 3D feature point cloud onto the 2D BEV plane using the camera's intrinsic and extrinsic parameters. Since the feature points generated from images can be quite dense, certain pixel features from the same image may project onto the same BEV grid. 
We utilize the BEV pooling operation to integrate all features on each BEV grid. Due to the computational complexity and the difficulties associated with deploying the BEV model, we adopt the BEV pooling optimization technique introduced in BEVFusion ~\cite{liu2023bevfusion} to simplify the model and meet the requirements of actual development. This approach accelerates the model training and reduces the inference time.

Compared to specific self-driving vehicles, most robots are usually equipped with a single front camera and move relatively slowly. 
Therefore, we define the BEV grid range as being located $20m$ in front and $5m$ behind the robot along the $x-$axis. The $x-$axis length of each grid cell is set to $0.25m$. Due to the narrower field of view in the $y-$axis when relying solely on input from the front-view camera, the $y-$axis range is defined as $-10m \sim +10$ of the robot, with a spacing of $0.2m$. 
The final BEV grid resolution is $100 \times 100$. In the depth estimation of BEV view transformation, We specify a depth range of $1m$ to $20m$ with an interval of $0.25m$, where each depth value corresponds to a pixel in the image.


\textbf{Task-driven Goal Decoder}: 
We design a navigation/exploration task-driven goal decoder based on the ViKiNG architecture ~\cite{shah2022viking}. 
For navigation tasks, the model takes the environmental context representation $o_{t-P}$, the current observation $o_{t}$, and the goal observation $o_{w}$ as input. 
It incorporates the BEV view transformation module described in the Local BEV perception encoder $p_\phi$ to improve the accuracy of latent features. 
In exploration tasks, the latent features are sampled from a Gaussian prior distribution $N(0, I)$. 
To ensure consistency of latent features between the navigation and exploration task, we adopt the VIB method~\cite{alemi2016deep} to train our objective function as follows
\begin{equation}
    \begin{split}
            \mathcal{L}_{VIB}(\theta,\phi)=E_\tau[-\mathbb{E}_{p\phi}[\log q_\theta(\{d\}_{t:t+H}|z^w_{t},o_{t-P:t}) \\
         + \lambda \log q_\theta(\{a,x\}_{t:t+H}|z^w_{t},o_{t-P:t})] \\
         + \beta \mathrm {KL}(p_\phi(z^w_{t}|o_{w},o_{t-P:t})||r(z^w_{t}))]
    \end{split}
    \label{eq_loss_vib}
\end{equation}
where $\tau$, $p_\phi$, $q_\theta$, $z^w_{t}$,  $d$, $a$, $x$, and $H$ represent a small batch of data per sample, the encoder,  the decoder, the latent features,  the temporal distance,  the waypoints, the GPS offset, and the prediction step size, respectively. $\lambda$ and $\beta$ are used to balance losses.

\begin{figure*}[thpb]
  \centering
  \includegraphics[scale=0.75]{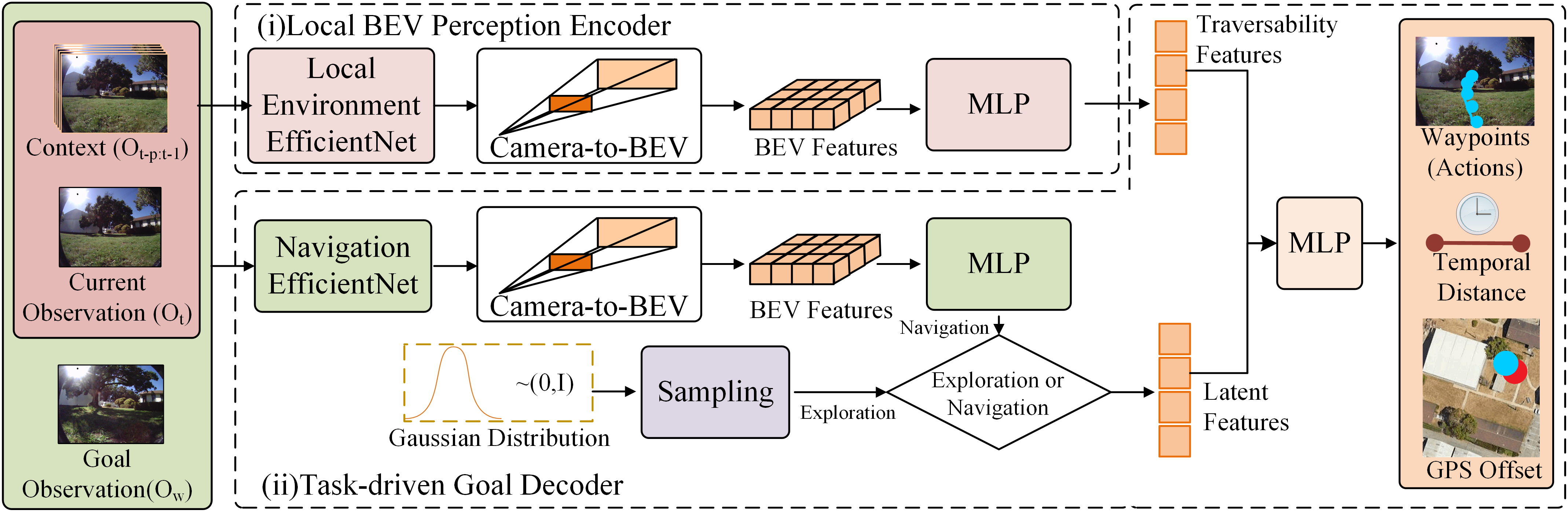}
  \caption{
  Task-driven goal decoder generates latent features based on task requirements, and by decoding the traversability feature and the latent feature, outputs the control information of the robot moving toward the sub-goal.
  }
  \label{LBPM}
\end{figure*}

The first two terms of Eq.~\ref{eq_loss_vib} are designed to establish a strong correlation in the \textit{LBPM}'s output, enabling the model to generate precise traversable waypoints.
This correlation is achieved by training with $\ell _{2}$ regression loss. 
The final term minimizes the KL divergence, ensuring the consistency of latent features under navigation/exploration tasks.
%
As a result, in the exploration task, the latent features sampled from the prior distribution can effectively generate waypoints, similar to those used in the navigation task.

\subsection{Global BEV Planning Model}
%
For long-distance path planning in unstructured outdoor environments, the robot faces challenges in identifying global traversability. 
Meanwhile, the various candidate traversable waypoints generated by the \textit{LBPM} require global hint information to select the optimal path.
Similar to structured navigation tasks, we intend to leverage environmental information to extract prior knowledge of given maps. 
Here, we discuss the form of the knowledge of maps, 
Human-annotated maps can guide robots well, but this leads to expensive labor costs. 
Common digital maps rarely reflect environmental landscapes such as trees and lawns, and they are more suitable for humans rather than robots. 
Overhead maps, as a form of large-scale BEV map representation, can effectively capture environmental elements. It provides an aerial view of surface conditions that can directly characterize terrain features with traversability.

To extract key traversable information from overhead maps, we have developed a model that learns traversability from overhead maps within the \textit{GBPM}. This approach effectively analyzes environmental elements and translates them into language that robots can easily comprehend. 
While overhead maps encompass various elements and can be segmented into distinct areas, such as houses and roads, this level of subdivision is unnecessary for outdoor navigation. We focus on traversability, guiding the robot to identify and select traversable areas.


\begin{figure}[thpb]
\centering
\subfloat[Overhead map]{
		\includegraphics[scale=0.35]{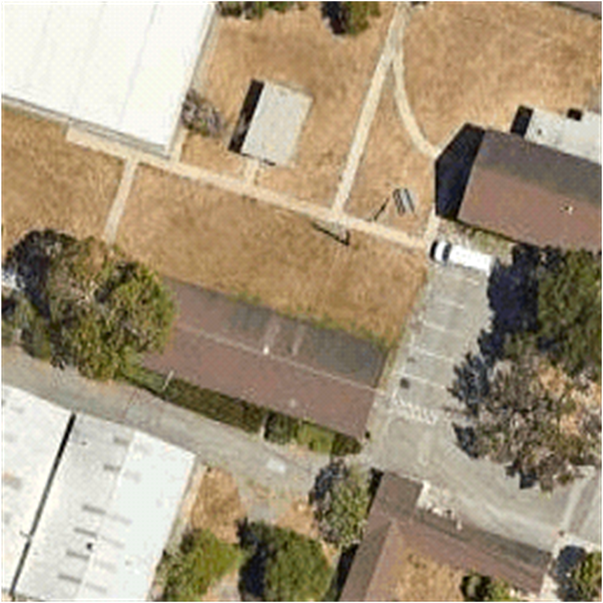}}
\subfloat[Other method]{
		\includegraphics[scale=0.35]{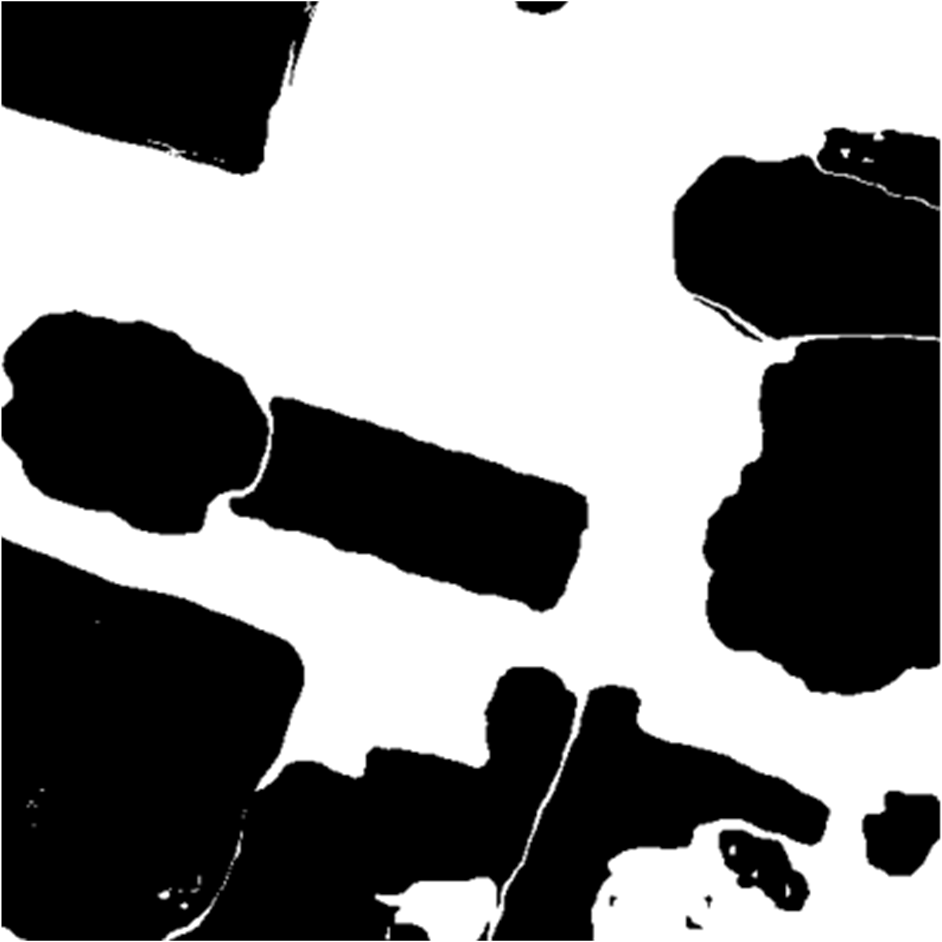}}
\subfloat[Our method]{
		\includegraphics[scale=0.35]{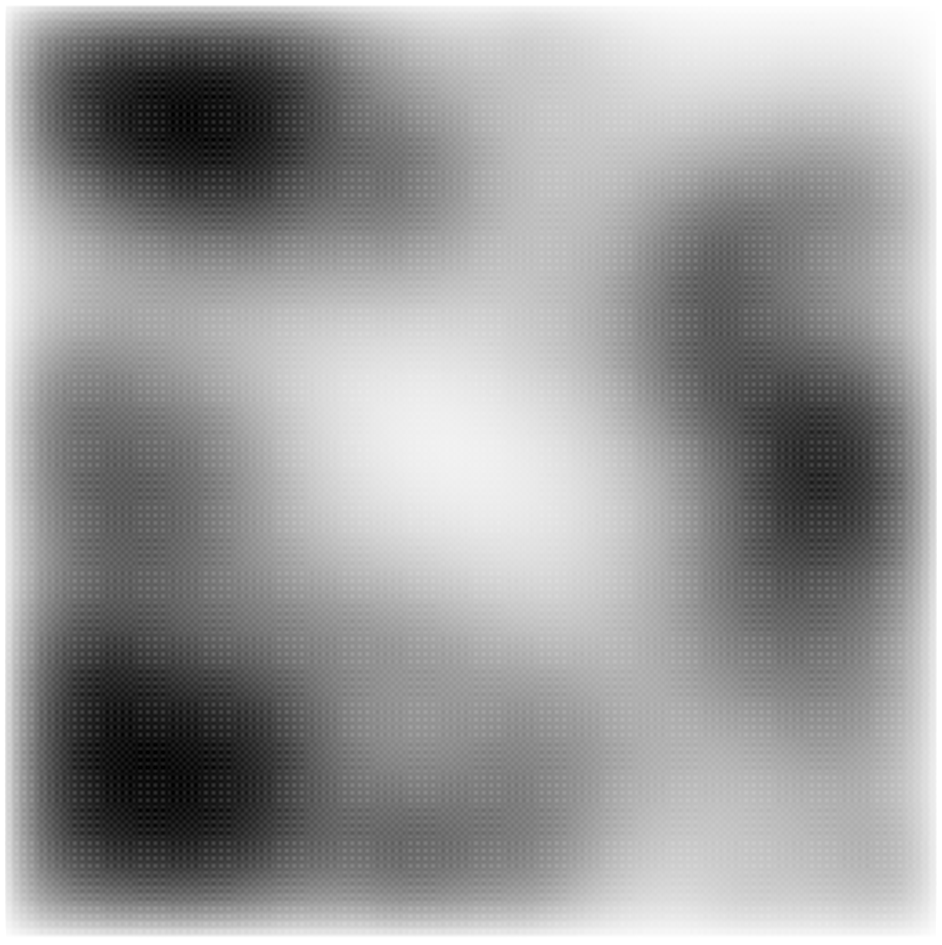}}
\caption{Two BEV feature extraction methods. (a): Original overhead map. (b): Precise segmentation of traversability. (c): Probability map of the traversability hints, generated from \textit{GBPM}.}
\label{GBPM-BEV}
\end{figure}

This approach differs from conventional segmentation and annotation, as shown in Figure \ref{GBPM-BEV}$(b)$. We avoid simple binary segmentation of the map, which can cause the robot to make an abrupt swerve and will only function when the robot is very close to the target. As shown in Figure \ref{GBPM-BEV}$(c)$, we build a probability map of traversability hints. 
In this map, values gradually increase as they approach impassable areas, allowing the robot to plan avoidance in advance. This continuous probability distribution is highly conducive to smooth and efficient robot navigation.

\textit{GBPM} uses trajectory data to learn traversability in the overhead map. In the trajectories generated by the robot’s autonomous exploration, the more easily accessible areas will be covered by a larger number of trajectories. This enables efficient traversability learning from trajectories. We utilize segmentation models to learn traversability. The historical path serves as the foreground, and the uncovered area serves as the background. The distinction is that we do not use thresholds to regulate the output, but directly use probability predictions as a map. U-Net~\cite{unet} is an outstanding segmentation model, which we employ to implement \textit{GBPM}.

\begin{figure}[thpb]
  \centering
  \includegraphics[scale=0.9]{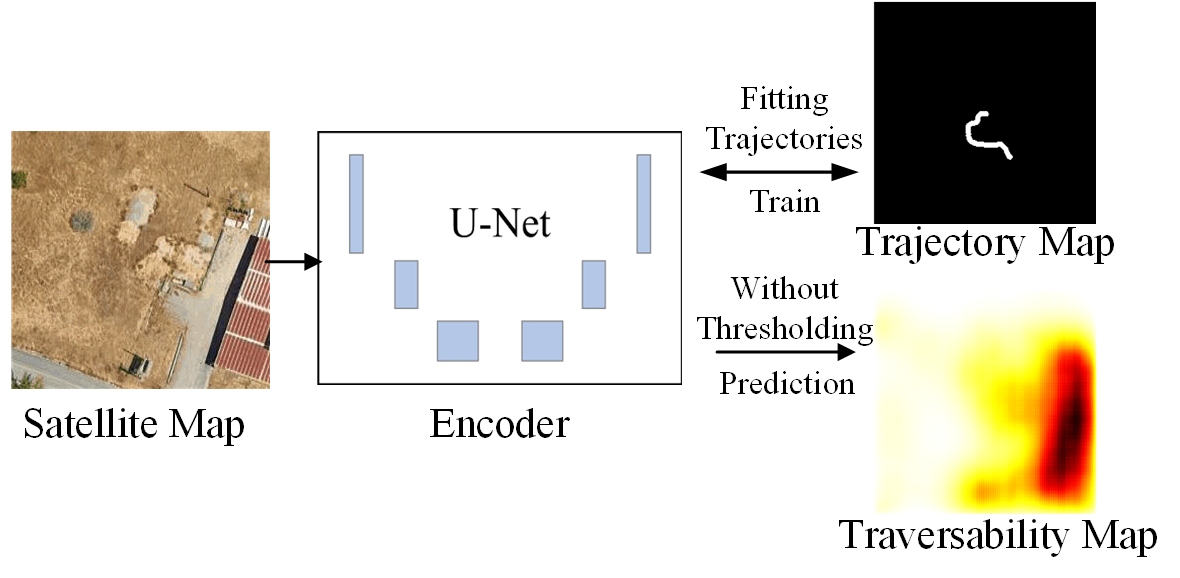}
  \caption{Training and inference architecture of \textit{GBPM}. The robot's trajectory is used as the ground truth to constrain during training. When making predictions, the probability map is directly used as hints.}
  \label{GBPM}
\end{figure}

To solve the problem of pixel imbalance between foreground and background, we use binary Focal Loss to train \textit{GBPM}. Its formula is the following Eq.~\ref{focal loss}:
\begin{equation}
    \text{Focal Loss} = -\alpha \cdot (1 - p_t)^\gamma \cdot \log(p_t)
    \label{focal loss}
\end{equation}
where $p_t$ represents the predicted probability of traversability obtained through the sigmoid function. The parameter $\alpha$ acts as a factor, while $\gamma$ adjusts the emphasis on well-classified examples and hard-misclassified examples. The training and inference schematic of the \textit{GBPM} model is shown in Figure \ref{GBPM}. 

\subsection{Integration of Local and Global BEV Planning}
We integrate local and global BEV planning models to enable long-range target-driven path planning. First, the \textit{LBPM} generates multiple potential traversable paths, providing information including temporal distance, GPS offsets, and waypoints from the current position to the goal. 
The \textit{GBPM} encodes the overhead map into a global probability map, projecting traversable paths of \textit{LBPM} onto this map. 
We can score the path and select them from a global view.
In detail, We evaluate the path using Eq.~\ref{total loss}, and the path with the minimum cost (i.e., maximum feasibility) is chosen as the optimal path.
\begin{equation}
    cost = k*score + (1-k)*temporal\_distance
    \label{total loss}
\end{equation}

In this formula, $k$ is a preset parameter, allowing the adjustment of the importance between \textit{GBPM} and \textit{LBPM}. Finally, the robot is guided toward its destination using the waypoint corresponding to the maximum feasibility path. This cycle repeats until the robot completes the task. The overall process is illustrated in Figure \ref{Overall architecture}.


\section{Experiments}
In this section, we conducted experiments on public datasets and real-world deployment, to compare the proposed \textit{LBPM} and \textit{GBPM} with baselines. 
\subsection{Experimental Settings}
We validated the proposed method using two ways: (i) temporal distance prediction via simulations on public datasets, and (ii) Real-world deployment experience.

For the simulations, we train the model using the RECON dataset~\cite{shah2021rapid}, which contains over $5,000$ trajectories from 9 different real-world scenarios. It includes onboard camera images, GPS coordinate information, and other dimensional features. The overhead image we used was obtained through the Google Maps API (developers.google.com/maps).

For the experiments on real-world scenarios, we took our school as the unstructured outdoor environment. This scene included a large number of grasslands, woodlands, and buildings. We chose a tree-shaded trajectory as the expected trajectory for the navigation experiment. Meanwhile, we selected several target points in the same scene for the signal-goal exploration experiment.

The comparison includes the ViKiNG~\cite{shah2022viking} and GNM~\cite{shah2023gnm} methods. 
In the preliminary work of the experiments, the RECON dataset is divided into a training set and a test set. Each of the three models is trained by a training set. 
The images from the original dataset are resized to dimensions of $ 128 \times 96 $pixels. 
We use the Adam optimizer~\cite{kingma2014adam} with a learning rate of $5e-4$ and a batch size of $32$ for training our models. For GNM training, we adhere to the GNM parameters and solely adjust the dimensions of the input images to preserve model performance as much as possible. 
As ViKiNG's work is not open-source, we use the model reproduced from its description. 
We train all models for $30$ epochs, and they converge within this period. 
Meanwhile, we used the test set to evaluate the models' temporal distance performance metrics after each training epoch, without adjusting any model parameters. 
The models were directly deployed on the robot for real-world comparative experiments. 
Two experimental tasks were designed: (i) a navigation comparison task over a distance of approximately $60m$, and (ii) a single-target exploration comparison task.
%

\subsection{Results and Analysis}
\textbf{Simulations on Public Dataset}: Temporal distance represents the time steps required to reach a sub-goal from the current position, reflecting the difficulty of reaching the sub-goal. It serves as an important metric for evaluating traversability. 
We validated the improvement of the proposed method by conducting comparative experiments on temporal distance performance metrics using the dataset:
(i) The accuracy of far/close goal classification: Given a target image, we classify its temporal distance from the current position. Specifically, goal images with a temporal distance ranging from 0 to 10 are labeled as "close," while those between 10 and 20 are labeled as "far." (ii) The accuracy of temporal-distance predictions: By evaluating the correctness of predictions under three conditions (difficult, medium, and easy). Using these metrics, we can evaluate the performance boundaries of different models in predicting traversability across various conditions.

As shown in Table \ref{Comparison of Accuracy}, methods utilizing context information (ours and GNM) for temporal distance prediction outperform those based solely on current observations (ViKiNG), demonstrating the crucial role of context in extracting traversability. Compared to the GNM method, our approach incorporates BEV view transformation, resulting in a $3\%$ improvement in the accuracy of far/close classification. In terms of the accuracy of temporal-distance predictions, we achieved significant improvements of $14.20\%$, $18.17\%$, and $17.75\%$ under error thresholds of $3$ (easy), $2$ (medium), and $1$ (difficult), respectively. This is attributed to the distance information and scale invariance in BEV, which enhances the robot's capability of identifying traversability and results in more accurate temporal-distance predictions.

\begin{table}[h]
    \centering
    \caption{Comparison of temporal distance prediction.}
    \begin{tabular}[t]{ccccc}
    \toprule 
          \multirow{2}{*}{\bfseries Method} 
        & \multirow{2}{*}{\begin{tabular}[c]{@{}c@{}}far\_or\_close\\($\%$)\end{tabular}} 
        & \multicolumn{3}{c}{$|$ dist\_pred - dist\_label $|$ $\leq error$
 ($\%$)} \\
        \cline{3-5} \addlinespace[0.5ex]
        &
        & $error=3$ 
        & $error=2$ 
        & $error=1$
        \\
    \midrule 
        ViKiNG & 92.30 & 62.92 & 49.25 & 30.82 \\
        GNM & 95.00 & 71.43 & 58.22 & 38.87 \\
        \bfseries \textit{LBPM} & \bfseries 97.77 & \bfseries 81.62 & \bfseries 68.85 & \bfseries 45.77 \\
    \bottomrule 
    \end{tabular}
    \label{Comparison of Accuracy}

\end{table}
 
\begin{figure}[thpb]
  \centering
  \subfloat[Waypoints prediction]{
		\includegraphics[scale=0.3]{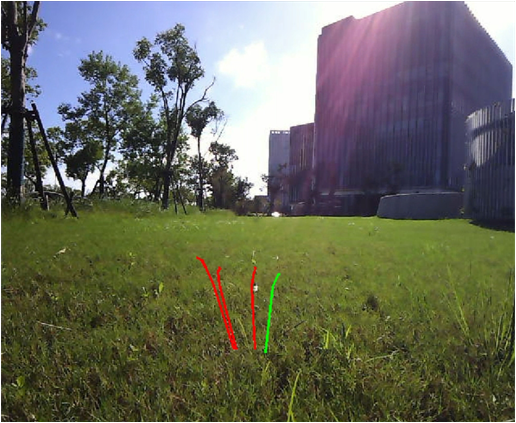}}
  \subfloat[Goal position observation]{
		\includegraphics[scale=0.3]{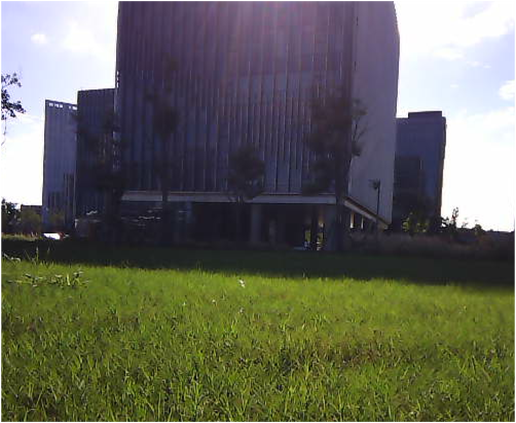}}
  \caption{Waypoints prediction visualization. (a): Waypoints prediction, the red line represents the available paths generated by the \textit{LBPM}, while the green line indicates the optimal waypoints selected based on the \textit{GBPM}. (b): Goal position observation. }
  \label{Action_visualization}
\end{figure}



\begin{figure}[thpb]
  \centering
  \includegraphics[scale=0.375]{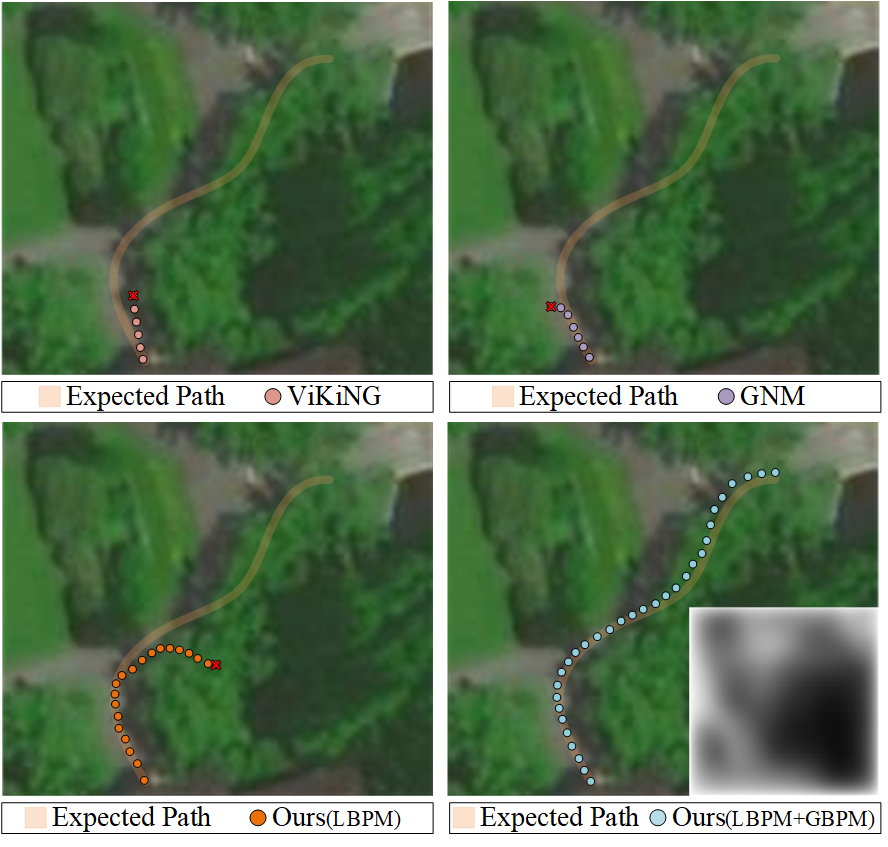}
  \caption{Comparison of different models on the navigation task. In ours (\textit{LBPM}+\textit{GBPM}), the image in the lower right corner shows the global probability map generated from the proposed \textit{GBPM}.}
  \label{path}
\end{figure}

\textbf{Real-world Deployment Experiments}: 
We compared the proposed approach with baselines (without global hints) in real-world environments by using a four-wheeled robot equipped only with a standard front camera and GNSS as sensors. 
(i) In the navigation task, despite significant differences between the test environment and the training dataset in terms of architectural style and surface vegetation, our method demonstrated remarkable robustness and achieved the best performance among all methods. It completed a 65-meter-long off-policy evaluation on a path with multiple sharp turns, as illustrated in Figure~\ref{path}.

Experimental results reveal that without global hints, ViKiNG, GNM, and \textit{LBPM} were unable to successfully complete the navigation task. 
ViKiNG and GNM demonstrated low capability in identifying traversability, with clear deviations from the expected path. 
While \textit{LBPM}, utilizing local BEV representation, can achieve navigation over certain distances, it inevitably suffers from path deviations and collisions due to the absence of global hints. 
The integration of \textit{LBPM} and \textit{GBPM} successfully completed the navigation task. 
The actual trajectory demonstrates that in areas without tree occlusions, the global traversability probability map from \textit{GBPM} provided valuable hints, effectively driving the robot. 
Even in scenes where tree occlusions caused global hints failure, the robustness and complementarity of the combined local and global BEV planning ensured the robot could adapt to environmental degradation and complete the task successfully.


(ii) To validate the boundary of our method for single-target exploration, we set targets at different distances. Targets were categorized into three levels: Easy (less than $10m$ from the current position), Medium ($10-20m$ from the current position), and Hard (more than $20m$ from the current position). The experimental results, as shown in Table \ref{Comparison of single-goal exploration task} and Table \ref{Comparison of displacement and velocity in Nsingle-goal exploration task}, reveal that in the absence of global hints, the proposed \textit{LBPM} outperforms baselines in hard and medium levels. 
With the addition of global hints, our method (\textit{LBPM}+\textit{GBPM}) achieves $100\%$ task completion at the Medium level and an $80\%$ success rate at the Hard level, with an average displacement of $31m$. 
This further demonstrates the enhancement in the capability of identifying traversability achieved by the integration of \textit{LBPM} and \textit{GBPM}.

\begin{table}[thpb]
    \centering
    \caption{Comparison of single-targe exploration task completion. 
    }
    \begin{tabular}[t]{cccc}
    \toprule 
          \multirow{2}{*}{\bfseries Method} 
        & \multirow{2}{*}{\begin{tabular}[c]{@{}c@{}}Easy\\$ \textless 10m$\end{tabular}} 
        & \multirow{2}{*}{\begin{tabular}[c]{@{}c@{}}Medium\\$10-20m$\end{tabular}} 
        & \multirow{2}{*}{\begin{tabular}[c]{@{}c@{}}Hard\\$ \textgreater 20m$\end{tabular}}\\

        \\
    \midrule 
        ViKiNG & 4/5  &  2/5&   0/5\\
        GNM & 5/5 & 3/5 & 1/5\\
        Ours(\textit{LBPM}) &  5/5 &  4/5 & 3/5\\
        \bfseries Ours(\textit{LBPM}+\textit{GBPM}) & \bfseries 5/5 & \bfseries 5/5 & \bfseries 4/5\\
    \bottomrule 
    \end{tabular}
    \label{Comparison of single-goal exploration task}
\end{table}

\begin{table}[h]
    \centering
    \caption{Comparison of average robot displacement and velocity in single-targe exploration task. }
    \begin{tabular}[t]{ccc}
    \toprule 
          \multirow{2}{*}{\bfseries Method} 
        & \multirow{2}{*}{Avg. Displacement($m$) } 
        & \multirow{2}{*}{Avg. Velocity($m\//s$)} \\

        \\
    \midrule 
        ViKiNG & 12 & 1.5  \\
        GNM & 17 & 1.5  \\
        Ours(\textit{LBPM}) &  22 &  1.5 \\
        \bfseries Ours(\textit{LBPM}+\textit{GBPM}) & \bfseries 31 & \bfseries 1.5 \\
    \bottomrule 
    \end{tabular}
    \label{Comparison of displacement and velocity in Nsingle-goal exploration task}
\end{table}



\addtolength{\textheight}{-4cm}   

\section{CONCLUSIONS}
In this work, we developed a method that combines local and global planning by integrating global BEV hints from overhead maps (e.g. satellite) with real-time local BEV representations.
This improvement enhances the robot's capability of identifying traversability and improves path planning distance in complex outdoor environments.
%
%
Due to factors such as changes in viewing angle of overhead maps, geographic information mismatches, and privacy concerns, obtaining overhead maps with high measurement accuracy in some scenarios remains challenging. 
In the future, we plan to incorporate unmanned aerial vehicles (UAVs) to obtain real-time aerial views as an upgrade to overhead maps. 
By integrating ground mobile robots and UAVs, the overall system's robustness and real-time performance 
are expected to be
significantly improved.




\clearpage


\bibliographystyle{IEEEtran}
\bibliography{arxiv}
 

\end{document}